# SHDB-AF: a Japanese Holter ECG database of atrial fibrillation

Kenta Tsutsui[1,2], MD, PhD; Shany Biton Brimer, PhD[3]; Noam Ben-Moshe, MSc[3,4]; Jean Marc Sellal, MD, PhD[5]; Julien Oster, PhD[6], Hitoshi Mori, MD, PhD[2], Yoshifumi Ikeda, MD, PhD[2], Takahide Arai, MD, PhD[2], Shintaro Nakano, MD, PhD[2], Ritsushi Kato, MD, PhD[2], and Joachim A. Behar, PhD[3]

Affiliations

1. Division of Cardiology, Department of Internal Medicine, Teikyo University School of Medicine
2. Department of Cardiology, Saitama Medical University International Medical Center
3. Faculty of Biomedical Engineering, Technion-IIT, Israel
4. Faculty of Computer Science, Technion-IIT, Israel
5. Department of Cardiology, Centre hospitalier Universitaire de Nancy, France
6. IADI UMR 1254, INSERM, Universite de Lorraine, Nancy, France

Correspondent

Kenta Tsutsui (knt22e@gmail.com)

## Abstract

Atrial fibrillation (AF) is a common atrial arrhythmia that impairs quality of life and causes embolic stroke, heart failure and other complications. Recent advancements in machine learning (ML) and deep learning (DL) have shown potential for enhancing diagnostic accuracy. It is essential for DL models to be robust and generalizable across variations in ethnicity, age, sex, and other factors. Although a number of ECG database have been made available to the research community, none includes a Japanese population sample. Saitama Heart Database Atrial Fibrillation (SHDB-AF) is a novel open-sourced Holter ECG database from Japan, containing data from 100 unique patients with paroxysmal AF. Each record in SHDB-AF is 24 hours long and sampled at 200 Hz, totaling 24 million seconds of ECG data.

## Background & Summary

Introduction

Cardiovascular diseases represent a significant threat to public health. AF is the most prevalent arrhythmia, it diminishes the quality of life and increases the risk of severe complications such as heart failure and stroke.[1,2] Because (1) maintaining sinus rhythm in patients with AF (so called "rhythm control") by either drugs or catheter ablation not only improves quality of life but also reverses structural abnormality of heart, and extends lifespan at least partly by reducing major cardiovascular events;[3-9] and (2) early treatment appears effective to enhance the treatment effect[10-13], early and acute diagnosis is crucial. However, current diagnostic tools for AF are suboptimal, leading to a lack of awareness and underdiagnosis of the condition.

ECG is the standard diagnostic tool of heart disease and required to confirm the diagnosis of AF. A challenge in AF diagnosis at its early stage is that a patient is in sinus rhythm for most time. In this context, diagnostic sensitivity crucially depends on the duration of ECG recordings, and across the modality, there is a trade-off between sensitivity and availability. The diagnostic sensitivity of standard 12-lead ECG is markedly hampered by brief recording times (10 seconds to 3 min) and limited access (only available at a healthcare provider facility). Holter ECG monitors are non-invasive and are capable of continuous monitoring <24 hours, playing a crucial role in a middle ground, and are therefore widely used as the de fact standard.[1,14-20] Besides such standard medical devices, we recently observe emergence of new health devices that are capable of ECG recordings such as implantable cardiac electronic devices and smart watches. Indeed, recent guidelines encourage use of such new devices for extended ECG monitoring.[1,19] Sophisticated software is necessary to maximize the diagnostic yield from the wide variety of inlets.

Prior works

There are a number of publicly available AF ECG database. These include the MIT-BIH arrhythmia database (MITDB) from USA including 48 two-channel ECG recordings of 30 minute each, sampled at 360 Hz with 11-bit resolution over a 10 mV range[21]. MIT-BIH Atrial Fibrillation database (AFDB) includes 23 long-term ECG recording with AF, mostly paroxysmal, 10 hours in duration, with two-channel ECG, each sampled at 250 samples per second with 12-bit resolution over a 10 mV range.[21,22] Recordings were manually annotated for AF, Atrial flutter (AFL) and AV junctional rhythms. Long Term AF database (LTAFDB) contains 84 long-term ECG recordings recorded by two-channels belonging to subjects with paroxysmal or sustained AF. Those recordings are of approximately 24 hours long digitized at 128 Hz with 12-bit resolution over 20 mV range.[23] Rhythms were automatically generated and manually verified by an experience team of ECG technicians.

More recently, IRIDIA-AF database,[24] including 167 paroxysmal AF tracings from 152 patients, 19-95 hours at 200Hz, emerged as a new resource for ML/DL research. Icentia11k is a database from Canada; 1-2 week-long continuous raw ECGs were recorded with single-lead CardioSTAT patch device, containing some AF events.[25] CPSC2021 contains total of 47 patients with AF events of varied duration from a single-lead device.[26]

Research gaps, objectives, and summary

A major challenge in translational medical AI is generalizability. Generalizability in DL refers to the ability of a model to perform well on new, previously unseen data that include a variety of geography, ethnicity, acquisition device, age groups and disease groups. Such databases are scarce. Especially, more data from north-east Asia (e.g., Japan, Korea, China and so on) is critically needed as the area is infamous for a rapid aging of the large population, leading to an ominous view that number of patients with age-related disorders including AF should markedly increase in near future.[27]

In the present work, we developed a new 24-hour Holter PAF database from Japan, a north-east Asian island. AF segment is manually annotated by certified cardiologists and with relevant clinical information.

## Methods and Technical Implementation

### Ethics

The present work was approved by the institutional ethics committee at Saitama Medical University International Medical Center (IRB number 2023-145). Due to retrospective and descriptive nature of the study, written informed consent was waived.

### Manual rhythm inspection and annotation

The PhysioZoo software[28,29] was used to annotate the Holter ECGs at the beat level. Supraventricular arrhythmias were divided and annotated into five categories: (1) AF, (2) AFL, (3) atrial tachycardia (AT), (4) Other supraventricular tachycardias such as Wolf-Parkinson-White and intranodal tachycardias and (5) other, such as NSR, that were not labeled. Reading time by the fellow (MA) was estimated to be on average 45 min per 24 h Holter recording.

### Data preparation

Holter recordings were recorded using Fukuda Holter monitor and digitized at 125 Hz from two available leads, CC5 and NASA, Figure 1. The recordings were then resampled to 200 Hz using an anti-aliasing filter as described in Biton et al.[30] Overall, 147 Holter recordings were collected between November 2019 and January 2022. Out of which a total of 100 recordings belonging to 100 unique patients were selected. The inclusion criteria for the selected recordings are described in Biton et al.[30] Essentially, those recordings were selected while stratified by age and sex. 80 recordings labeled as $AF_l$ and 20 labelled as non-$AF_l$ were randomly selected based on the medical reports which were prepared following the patient's examination. $AF_l$ is a combined label for AF and AFL.

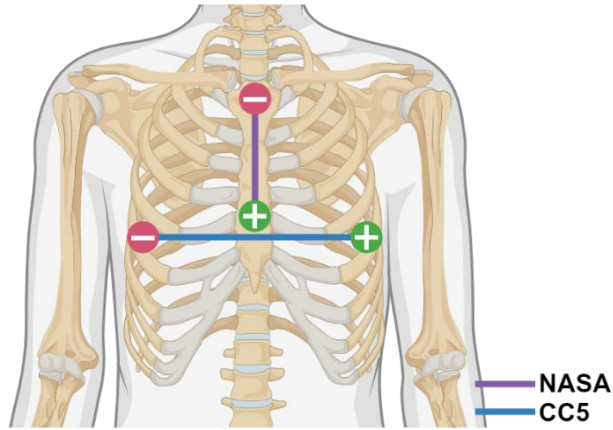

*Figure 1: Electrode placement of Holter leads used in this research.*

## Data Records

<u>Patient characteristics</u>

Table 1 summarizes patient characteristics of the SHDB AF database.

The AF burden (AFB) is defined as the percentage time spent in AF, equation 1.

$$AFB = \frac{\sum_{n=1}^{N} t_i \times \mathbb{1}_i}{\sum_{n=1}^{N} t_i} \qquad 1$$

N represents the number of available windows, $t_i$ represents the length of the $i_{th}$ window in milliseconds and $\mathbb{1}_i$ is equal to 1 for AF windows and zero otherwise.

The AFB is shown in Figure 2, with mean $\pm$ std of $19.5 \pm 27.6\%$.

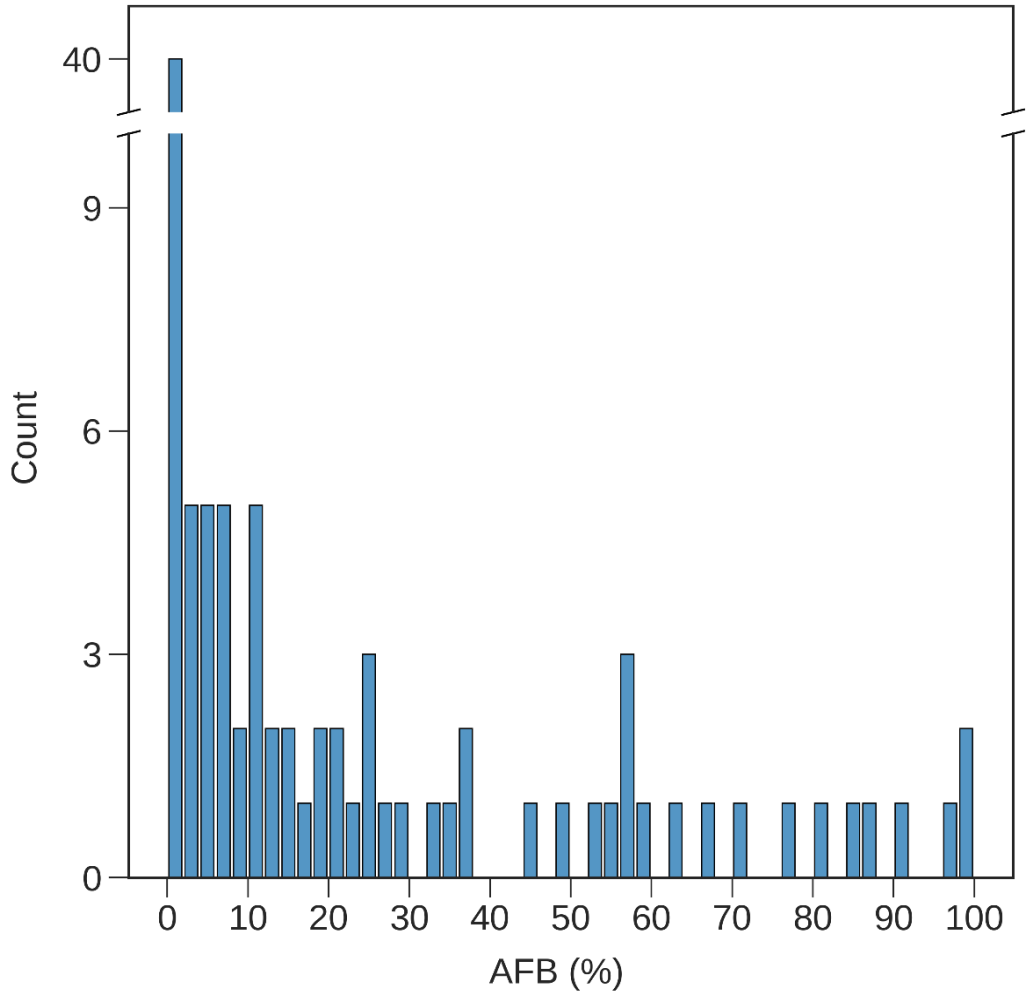

*Figure 2: Histogram of the AFB (AF Burden) for SHDB (n=100).*

Database description

The data files are provided in open WFDB standard format. The ECG waveforms are stored in *.dat files. The header files (.hea) are associated with specific recording files and specifies attributes such as associated .dat file name, number of leads, sampling frequency, recording date and time. The .qrs files are the R-peak annotation files detected by epltd algorithm[31] and were used as the input data to an DL algorithm for AF detection describe in the Performance section. ECG example is given in

Figure 3.

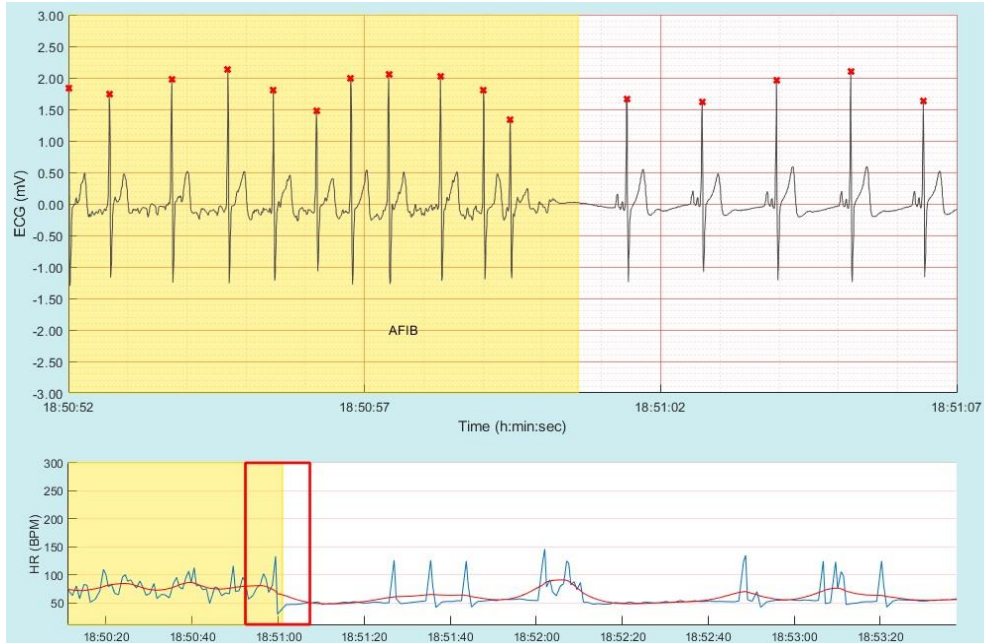

*Figure 3: Segment of an ECG example extracted as part of this research shown using PhysioZoo [28], AFIB: Atrial Fibrillation.*

Technical Validation Performance

Signal quality index (bSQI)[32] was computed for non-overlapping windows of 60-beat. Reference beat annotation were detected using the epltd implementation of the Pan and Tompkins algorithm.[31] Those were compared against test peaks detected with xqrs[33], with an agreement window of 50ms. The median and interquartile range (Q1-Q3) for 60-beat windows was 1.0 (1.0-1.0) and 0.999 (0.997-0.999) for the entire recordings.

We evaluated a DL algorithm for AF event detection from beat-to-beat intervals termed ArNet2[30] as part of an attempt to show its generalizability on different distribution shifts. ArNet2 accepts as an input 60-beat RR-intervals and outputs a binary classification of $AF_l$/non-$AF_l$ per window. It is divided into two parts; first part is a combination of residual blocks used to build representation of an ECG signal window. The second part consists of Gated Recurrent Units (GRU) which learns from the temporal dependencies in time series. ArNet2 was trained on a subset of UVFDB and evaluated on 4 different databases from multiple countries and on a combined test set combining all test set together to assess global performance. A second DL named RawECGNet[34] was developed and benchmarked against ArNet2. As opposed to ArNet2, RawECGNet was developed based on the raw, single-lead ECG. The algorithm was trained and evaluated on the same databases as ArNet2, however the input windows are 30-seconds long of the raw ECG. In addition to the traditional evaluation metrics, the AFB was estimated. Looking at F1 score, ArNet2 achieved F1 of 0.92 for SHDB using the beat-to-beat time intervals and 0.93 for RawECGNet on the raw ECG signal. Further results are presented at Biton et al.[30] and Ben-Moshe et al[34].

## Code Availability

Physiozoo software is available at https://physiozoo.com

## Acknowledgments

Hittman: Technion EVPR Fund: Hittman Family Fund.

## Author Contributions

KT conceptualized the project, gathered, and managed clinical data associated with the present study, and drafted the manuscript. KT, JMS, JO read and annotated Holter ECG. SBB, NBM, JAB performed computational work, prepared the database for open release and revised the manuscript. HM, YI, TA, SN and RK reviewed and revised the manuscript.

## Competing interests

None declared.

| | | |
|---|---|---|
| n | | 100 |
| Age, mean (SD) | | 68.0 (11.3) |
| Female, n (%) | | 45 (45.0) |
| Height (m), mean (SD) | | 1.6 (0.1) |
| Weight (kg), mean (SD) | | 61.1 (14.4) |
| BMI, mean (SD) | | 22.9 (4.2) |
| AF type, n (%) | Paroxysmal | 69 (69.0) |
| | Persistent | 11 (11.0) |
| | Sinus rhythm (non-AF) | 20 (20.0) |
| History of AFL, n (%) | | 14 (14.0) |
| Previous ablation, n (%) | | 28 (28.0) |
| Pacemaker, n (%) | | 3 (3.0) |
| Antiarrhythmic drugs, n (%) | Bepridil | 3 (3.0) |
| | Pilsicainide | 1 (1.0) |
| | Amiodarone | 10 (10.0) |
| | Cibenzoline | 2 (2.0) |
| | Frecainide | 1 (1.0) |
| | Pilsicainide | 3 (3.0) |

*Table 1: Patient characteristics of the SHDB-AF dataset.*